\pdfoutput=1
\documentclass[11pt]{article}

\usepackage{hyperref}
\usepackage{natbib}
\usepackage{subfiles}

\usepackage{acl}
\usepackage{booktabs}

\usepackage{times}
\usepackage{latexsym}

\usepackage{amsmath}
\usepackage{amsfonts}
\usepackage{amstext}
\usepackage{upgreek}
\usepackage{graphicx}  
\usepackage{dsfont}
\graphicspath{ {./images/} }

\usepackage{multirow}
\usepackage{tabu}
\usepackage{xcolor}
\usepackage{subcaption}
\usepackage{caption}
\usepackage[T1]{fontenc}

\usepackage[utf8]{inputenc}

\usepackage{microtype}

\usepackage[all]{nowidow}

\newcommand{\STD}[1]{\scriptsize$\scriptscriptstyle\pm$#1}

\def\data{D}
\def\errorSet{E}
\def\dataSize{N}
\def\dataset{\data={\{(x_i,y_i)\}_{i=1}^{\dataSize}}}
\def\emb{\textbf{\textrm{e}}}
\def\penu{\textbf{\textrm{x}}}
\def\cov{\boldsymbol{\Sigma}}
\def\mean{\boldsymbol{\mu}}
\def\mahalEq{M(\penu) = \sqrt{(\penu-\mean_y)^\top\cov_y^{-1}(\penu-\mean_y)}}

\def\ERM{\textbf{\textrm{ERM}}}
\def\Jtt{\textsc{Jtt}}
\def\JTT{\textbf{\Jtt}}
\def\Jttm{\Jtt\text{-m}}
\def\JTTm{{\textbf{\JTT-m}}}

\title{Outlier-Aware Training for Improving Group Accuracy Disparities}

\author{
Li-Kuang Chen$^1$\thanks{$\;\;$This work was conducted during the author's internship under National Institute of Informatics, Japan.} \hspace{2em} Canasai Kruengkrai$^2$ \hspace{2em} Junichi Yamagishi$^2$ \\
$^1$National Tsing Hua University, Taiwan \\
\texttt{lkchen@nlplab.cc} \\
$^2$National Institute of Informatics, Japan \\
\texttt{\{canasai,jyamagishi\}@nii.ac.jp} }

\begin{document}
\maketitle
\begin{abstract}
Methods addressing spurious correlations such as Just Train Twice ($\Jtt$,~\citealt{liu2021just}) involve reweighting a subset of the training set to maximize the worst-group accuracy. However, the reweighted set of examples may potentially contain unlearnable examples that hamper the model's learning. We propose mitigating this by detecting outliers to the training set and removing them before reweighting. Our experiments show that our method achieves competitive or better accuracy compared with $\Jtt$ and can detect and remove annotation errors in the subset being reweighted in \Jtt.\footnote{Our code is available at \url{https://github.com/nii-yamagishilab/jtt-m}.}
\end{abstract}

\section{Introduction}

Machine learning models trained with empirical risk minimization (ERM,~\citealt{vapnik92}) can achieve a high average accuracy by minimizing the overall loss during training. 
Despite this, ERM models are also known to perform poorly on certain minority groups of examples. 
When specific attributes in a dataset frequently co-occur with a class label, ERM models often learn to correlate the co-occurring attributes and the label, using the attributes as ``shortcuts'' for classifying examples. 
These ``shortcuts'' are also called {\em spurious correlations}, because model performance can significantly decrease when the model encounters examples that belong to a minority group where the correlations between the attributes and class label do not hold. 

\begin{figure}[t]
\centering
\includegraphics[width=0.95\columnwidth]{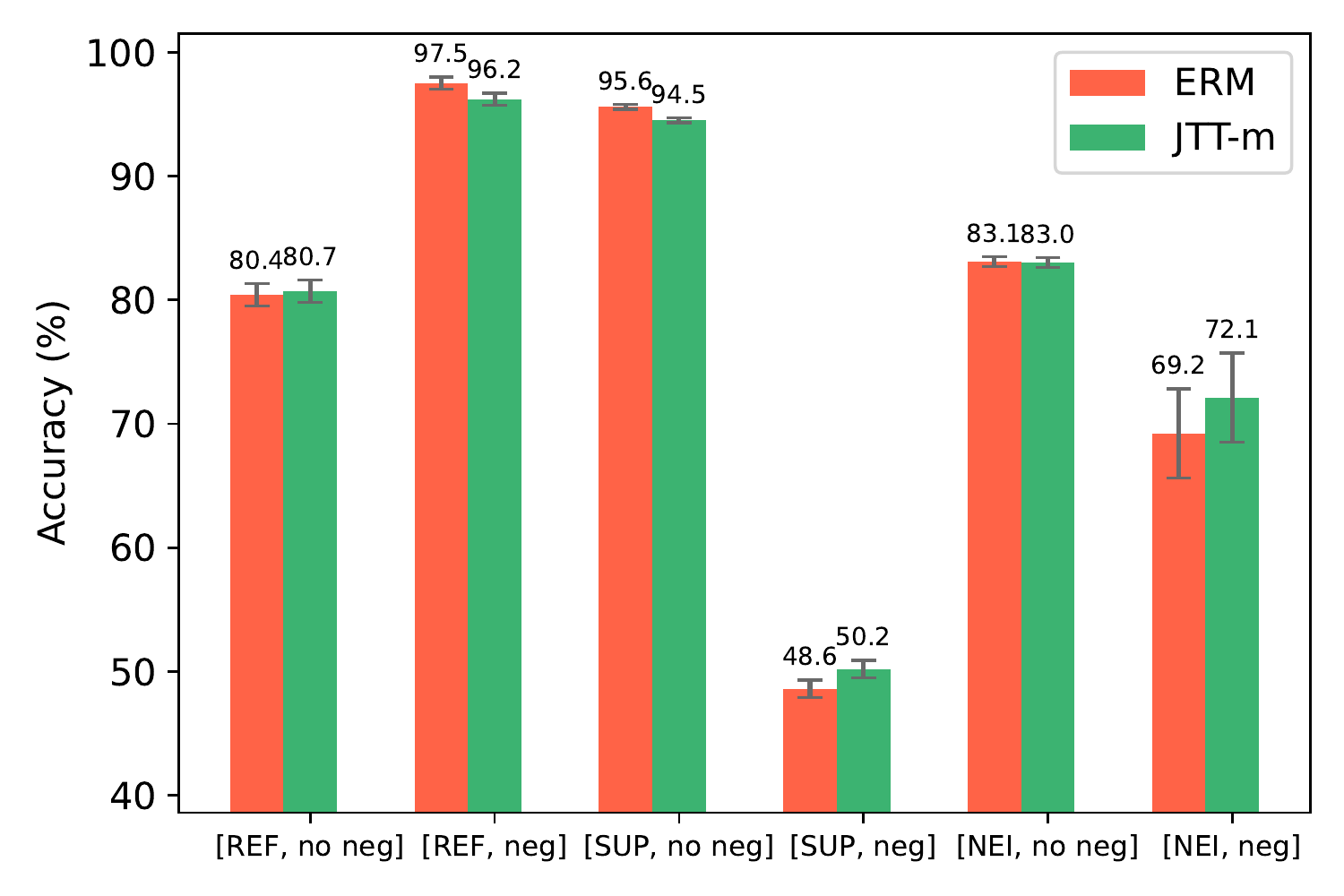}
\caption{
Results for the FEVER test set~\cite{thorne2018fever,schuster2021vitc}.
The data are divided into six groups in accordance with  {\em class-attribute} combinations, where class = \{\textsc{Refutes} (REF), \textsc{Supports} (SUP), \textsc{Not Enough Info} (NEI)\} and attribute = \{no neg, neg\}, indicating the presence of a negation word in the claim.
Both methods perform well on groups with strong spurious correlations (e.g., [REF, neg]).
Our proposed method ($\Jttm$) helps improve accuracies for groups where such spurious correlations do not appear (e.g., [SUP, neg] and [NEI, neg]).
}
\label{fig:FEVER-test-acc}
\end{figure}

More specifically, each class in a dataset can be divided by whether their examples contain such spurious attributes. Each set of examples with a class-attribute combination is called a ``group''. The worst group is characterized by having the poorest ERM model performance among other groups. 
As an example, Figure~\ref{fig:FEVER-test-acc} shows accuracy disparities among groups in the FEVER dataset.
The ERM-trained model can achieve close to perfect accuracy on the group with a spurious correlation (the \textsc{Refutes} class with negation), but only half the accuracy on the worst group (the \textsc{Supports} class with negation). 
  
Improving the worst-group performance of ERM models while maintaining the overall accuracy is an active topic of research that has applications in fair machine learning classifiers or robustness against adversarial examples~\cite{slowik2022distributionally}. 
Methods aiming to maximize worst-group accuracy can be roughly categorized into two categories: those that utilize group information and those that do not. 
Group Distributionally Robust Optimization (Group DRO,~\citealt{sagawa2019distributionally}) uses attribute (and thus group) information during training to dynamically minimize the loss of each group. 
While Group DRO achieves a high worst-group and overall accuracy, it requires annotation on group information during training, which can be expensive to obtain and unavailable for less popular datasets. 
On the other hand, methods such as DRO with Conditional Value-at-Risk (CVaR DRO,~\citealt{duchi2020distributionally,levy2020large}), Learning from Failure (LfF,~\citealt{nam2020learning}), Predict then Interpolate (PI,~\citealt{bao2021predict}), Spectral Decoupling (SD,~\citealt{pezeshki2021gradient}), Just Train Twice ($\Jtt$,~\citealt{liu2021just}), and RWY and SUBY from \cite{idrissi2022simple} all aim to minimize worst-group loss without group information.

CVaR DRO minimizes worst-case loss over all subpopulations of a specific size and requires computing the worst-case loss at each step. 
LfF trains an intentionally biased model and upweights the minority examples. 
PI interpolates distributions of correct and incorrect predictions and can minimize worst-case loss over all interpolations. 
SD replaces the L$_2$ weight decay in the cross entropy loss function with logits. 
RWY reweights sampling probabilities so that mini-batches are class-balanced. SUBY subsamples large classes so that every class is the same size as the smallest class.
$\Jtt$ simply obtains misclassified examples (the \emph{error set}) from the training set once and upweights the fixed set of erroneous examples. 
We focus on $\Jtt$ due to its simplicity and relative effectiveness and because it does not require group information for improving worst-group accuracy. While \citet{idrissi2022simple}'s SUBY and RWY also follow $\Jtt$ in improving worst-group accuracies, their methods target only datasets with imbalanced classes, and are not applicable to class-balanced datasets such as MultiNLI~\cite{williams2018broad}.
    
We propose further enhancing $\Jtt$ by removing outliers from the error set before upweighting it. 
The outliers might be examples that are difficult to learn, such as annotation errors. 
Keeping them from being upweighted allows the model to train on a cleaner error set and thus better show the intended effect of the original $\Jtt$.
We focus on worst-group performance caused by the spurious correlations of negation words and evaluate on datasets susceptible to spurious correlations of this type. 
Our experiments on the FEVER and MultiNLI datasets show that our method can outperform $\Jtt$ in terms of either the average or the worst-group accuracy while maintaining the same level of performance for the other groups.

Our contributions are as follows. 
We devise a method for improving worst-group accuracy without group information during training based on $\Jtt$ (Section \ref{sec:proposed-method}). 
We show that by removing outliers from the error set being upweighted, we can achieve similar or better overall and worst-group performance (Section \ref{subsec:results}). 
Our examination of the outliers being removed also suggests that the improvement may come from removing annotation errors in the upweighted error set (Section \ref{subsec:discussion}).

\section{Background}

\subsection*{Spurious correlations and minority groups} 

We investigate the spurious correlations occurring in two natural-language datasets: FEVER~\cite{thorne2018fever} and MultiNLI~\cite{williams2018broad}. 
The task for FEVER involves retrieving documents related to a given {\em claim}, finding sentences to form {\em evidence} against the claim, and then classifying the claim on the basis of the evidence into three classes: \textsc{Supports} (SUP), \textsc{Refutes} (REF), or \textsc{Not Enough Information} (NEI).
We focus on improving the worst-group classification performance for the final part of the task. 
The task for MultiNLI is to classify whether the \emph{hypothesis} is entailed by, neutral with, or contradicted by the \emph{premise}. 
We use \citet{schuster2021vitc}'s preprocessing of both datasets, containing 178,059/11,620/11,710 training/dev/test examples for FEVER and 392,702/9,832 training/test examples for MultiNLI.

Attributes known to cause spurious correlations for these datasets are negation words \cite{gururangan2018annotation} and verbs that suggest negating actions \cite{schuster2019towards}. 
We merge these two sources of negation words into a single set: \{\emph{no, never, nothing, nobody, not, yet, refuse, refuses, refused, fail, fails, failed, only, incapable, unable, neither, none}\}.
Each class can be split into two groups based on whether each claim/hypothesis contains a spurious attribute (i.e., the negation words listed above). 
Models tend to perform well on groups where the attributes are highly correlated with the label. 
Groups where the correlation between the label and the attribute does not hold are called {\em minority groups} or {\em worst groups}, since models often fail to classify their examples correctly. 
For example, the claim ``\emph{Luis Fonsi does \textbf{not} go by his given name on stage.}'', labeled \textsc{Supports}, belongs to the worst group [SUP, neg].

Table \ref{tab:class-group-distributions}(a) shows that most claims containing negation are from the class \textsc{Refutes}. 
The relatively small amount of examples from the groups (SUP, negation) and (NEI, negation) form the minority groups, where the ERM model performance fails. 
A similar trend can be seen in Table \ref{tab:class-group-distributions}(b).

\begin{table} [t]
\small
    \begin{subtable}[ht]{\columnwidth}
    \setlength{\tabcolsep}{2.5pt} 
    \begin{center}
    \begin{tabu}{lccc} 
    \toprule
                & No negation &  Negation &  \\
    \cmidrule{2-3}
    REF &  27,575 (17.1\%) & 14,275 (86.3\%)                         &  \phantom{1}41,850 (23.5\%) \\
    SUP &  99,303 (61.5\%) & \phantom{1}1,267 \phantom{ } (7.7\%)    &  100,570 (56.5\%) \\
    NEI &  34,633 (21.4\%) & \phantom{1}1,006 \phantom{ } (6.0\%)    &  \phantom{1}35,639 (20.0\%) \\
    \toprule
    \end{tabu}
    \caption{FEVER}\bigskip\label{sub-tab:class-grp-a}
    \end{center}
    \end{subtable}
    \begin{subtable}[ht]{\columnwidth}
    \setlength{\tabcolsep}{2.5pt} 
    \begin{center}
    \begin{tabu}{lccc} 
    \toprule
                & No negation &  Negation &  \\
    \cmidrule{2-3}
    Contr &  \phantom{1}88,180  (27.3\%)  & 42,723 (61.2\%) & 130,903 (33.3\%) \\
    Ent   &  118,554 (36.7\%)  & 12,345 (17.7\%) & 130,899 (33.3\%) \\
    Neut  &  116,185 (36.0\%)  & 14,715 (21.1\%) & 130,900 (33.3\%) \\
    \toprule
    \end{tabu}
    \caption{MultiNLI}\label{sub-tab:class-grp-b}
    \end{center}
    \end{subtable}
\caption{
Class and group distributions for (a) FEVER and (b) MultiNLI training sets. Both datasets show a high spurious correlation between the REF (Contr) class and the attribute neg. Minority groups where the spurious correlation does not hold are [SUP (Ent), neg] and [NEI (Neut), neg].
}\label{tab:class-group-distributions}
\end{table}


\subsection*{Empirical Risk Minimization (ERM)}
  
Let $x \in \mathcal{X}$ be a training example and $y \in \mathcal{Y}$ be its label.
Given a dataset $\dataset$, ERM aims to minimize the average loss (``empirical risk''), defined as:
\begin{equation}\label{eqn:erm}
    J_{\rm ERM}(\theta) = \frac{1}{\dataSize}\sum_{(x,y) \in \data}\ell(g_\theta(x), y),
\end{equation}
 where $\dataSize$ is the number of training examples, $g_\theta(\cdot)$ is the model, and $\theta$ represents model parameters.
We use cross-entropy loss as the loss function:
\begin{equation}
    \ell(g_\theta(x), y) = -\sum_{y\in\mathcal{Y}} \mathds{1} \{y=\hat{y}\} \text{log}(p_\theta(\hat{y}|x)),
\end{equation}
where $\mathds{1}\{\cdot\}$ is the indicator function, $x$ represents the input sentence pair ($s_1, s_2$),  and $y\in\mathcal{Y} = $\{SUP, REF, NEI\} (\{Ent, Contr, Neut\} for MultiNLI).
We first encode the input sentence pairs with BERT~\cite{devlin2018bert} and feed the resulting embedding $\emb$ into a multi-layer perceptron (MLP) followed by a softmax function for classification:
\begin{equation}
    \begin{split}
    p_\theta(\hat{y}|x) &= \text{softmax}(\text{MLP}(\emb)),\\
    \emb &= \text{BERT}(s_1, s_2).
    \end{split}
\end{equation}

\subsection*{Just Train Twice ($\Jtt$)}

\newcite{liu2021just} propose improving worst-group performance by simply training with an upweighted error set. 
During the first round of training, the set of incorrectly classified training examples is identified via an ERM model. 
The training error set $\errorSet$ is then upweighted with a real and positive upweight factor $\lambda_{\rm up} \in \mathbb{R}^+$, and a final model is trained on the reweighted objective:

  \begin{equation}\label{eqn:jtt}
\setlength{\medmuskip}{0mu}
  \begin{split}
        J_{\rm up}&(\theta, \errorSet) \\
        = 
         &\frac{1}{N_{\rm up}} 
        \biggl(
        \lambda_{\rm up} \sum_{\substack{(x, y) \\ \in \errorSet}} \ell (g_\theta(x), y) 
        + \sum_{\substack{ (x,y) \\ \notin \errorSet}} \ell(g_\theta(x), y)
        \biggr)
        ,
  \end{split}
  \end{equation}
  where  $\lambda_{\rm up}$ is a hyperparameter, and $N_{\rm up}$ is the size of the training set after upweighting.

\section{Proposed method} \label{sec:proposed-method}

Even though the upweighted ERM error set can improve worst-group accuracy, it is possible that the error set contains {\em unlearnable} or {\em out-of-distribution} (OOD) examples, e.g., annotation errors. When upweighting the entire error set, these examples will get amplified along with the rest of the error set, lessening the overall benefits of upweighting and retraining.


We propose modifying the $\Jtt$ algorithm by removing outliers in the ERM error set before training the second time. 
We adopt a similar approach from \citet{lee2018simple} for detecting outliers.
Let $\penu$ be the output of the penultimate layer (i.e., the last layer before the logits) and belong to class $y$.
First, we calculate the Mahalanobis distance for each $\penu$ from the mean of each class $y$:
\begin{equation}
\mahalEq,
\end{equation}
where $\mean_y$ and $\cov_y$ are the class mean and covariance.\footnote{We compute $\cov_y$ using the standard covariance maximum likelihood estimate (MLE) implemented in scikit-learn.}
The greater the distance of $\penu$ is from $\mean_y$, the likelier it is to be an OOD example. 

\begin{figure}[t]
\centering
\includegraphics[width=1.0\columnwidth]{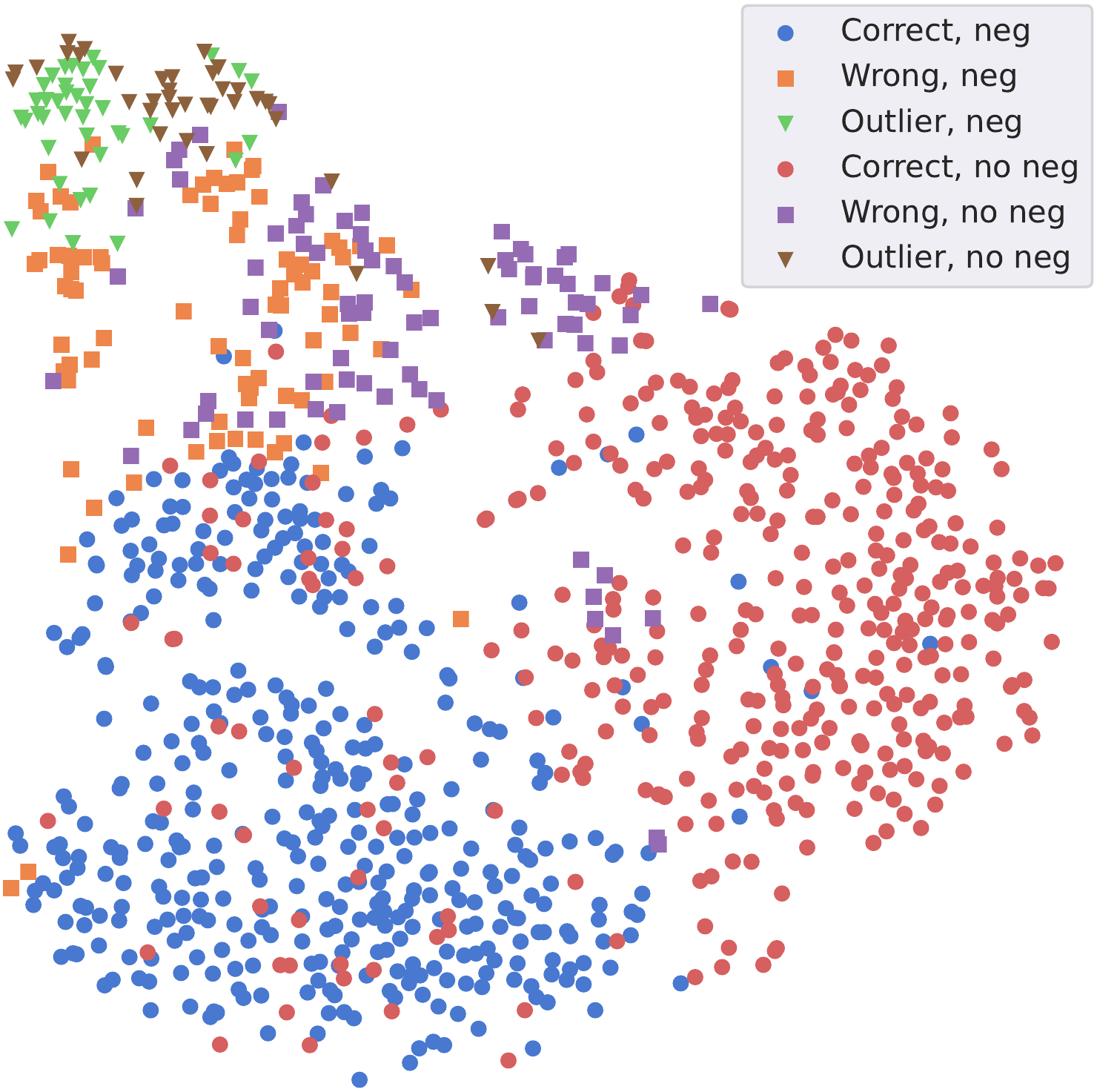}
\caption{
T-SNE visualization of samples from the class Entailment of the MultiNLI training set. 
Correct predictions of groups with and without negation (blue and red) are quite well separated. Wrong predictions lie at the top left, and outliers lie further away. 
Outliers are detected by their Mahalanobis scores.
}
\label{fig:MNLI-tsne}
\end{figure}

Then, we filter OOD examples by comparing the calculated Mahalanobis distance against a chi-squared distribution with a critical value $\alpha$ of 0.001 and a degree of freedom $df$:\footnote{We select a value of $df$ that yields the best worst-group accuracy on the dev set.}
\begin{equation} \label{eqn:ood-threshold}
 x_i \in 
\begin{cases}
  S_{\rm in} & \mathrm{if} \; \; \;
  p_i < \alpha, \\
  S_{\rm out} & \mathrm{if} \; \; \;
  p_i \geq \alpha, \\
\end{cases}
\end{equation}
where $S_{\rm in}$ and $S_{\rm out}$ are the sets of in-distribution and OOD training examples, and $p_i$ is the $p$-value of the $i$-th example.
We show the T-SNE visualization in Figure~\ref{fig:MNLI-tsne}.

\begin{table*}[t]
    \centering
    \begin{tabular}{lcccc} 
    \toprule
    \multicolumn{1}{l}{Dataset}      & 
    \multicolumn{2}{c}{FEVER}       & 
    \multicolumn{2}{c}{MultiNLI}   \\
    \cmidrule(lr){2-5}
                                &
     Avg. (\%) & Worst (\%)    & 
     Avg. (\%) & Worst (\%)    \\    
    \midrule
     ERM  & 87.8\STD{0.2}$\;\;$ & 48.6\STD{0.7} & 84.9\STD{0.1} & 72.0\STD{1.0}$\;\;$ \\
     $\Jtt$  & 86.8\STD{0.2}$\;\;$ & 50.5\STD{3.5} & 83.0\STD{0.2} & 75.5\STD{1.5}$\;\;$ \\
     $\Jttm$ &	87.4\STD{0.1}$^\ast$ & 50.2\STD{2.8} & 83.0\STD{0.3} & 77.3\STD{0.4}$^\ast$ \\
    \bottomrule
    
    \end{tabular}
    \caption{Average and worst-group test accuracies for all methods. The ``Worst'' column indicates the worst-group accuracies on [SUP, neg] and [Neutr, neg] for FEVER and MultiNLI, respectively. We report mean and standard deviation computed across five runs using different random seeds. ``$\ast$'' indicates the statistical significance compared with $\Jtt$  (paired t-test, $p < 0.05$).}\label{tab:main-results}
\end{table*}

Once the OOD examples are identified, we remove the subset of misclassified OOD examples from the error set $E$, forming a new error set $E_{\rm in}$:
\begin{equation}\label{eqn:remove-ood}
E_{\rm in} = \{(x_i, y_i) \; \text{s.t.} \; 
\hat{y_i} \neq y_i \land  x_i \notin S_{\rm out})\},
\end{equation}
which is then upweighted as per $\Jtt$:

\begin{equation}\label{eqn:jtt-m}
\setlength{\medmuskip}{0mu}
  \begin{split}
        &J_{\rm up\mbox{-}in}(\theta, \errorSet_{\rm in}) \\
        &= 
        \frac{1}{N_{\rm up}} 
        \biggl(
        \lambda_{\rm up} \sum_{\substack{(x, y)\\\in \errorSet_{\rm in}}} \ell (g_\theta(x), y) 
        + \sum_{\substack{ (x,y) \\ \notin \errorSet_{\rm in}}} \ell(g_\theta(x), y)
        \biggr)
        ,
  \end{split}
  \end{equation}

\section{Experiments}

\subsection{Training details}

We follow~\newcite{sagawa2019distributionally, liu2021just, idrissi2022simple} in using different optimization settings for different training methods to maximize the validation accuracy.
For ERM, we used the AdamW optimizer \cite{Loshchilov2019DecoupledWD}, linear learning rate decay, and a gradient clipping of 1. 
For the first training of $\Jtt$, we used the SGD optimizer without gradient clipping. 
The second training used the same settings as those of ERM.

We used HuggingFace's implementation \cite{wolf2020transformers} of BERT-base with default parameter settings.
For all methods, we used a batch size of 32, initial learning rate of 2e-5, and we trained them for 2 epochs. 
We tried $df \in \{4, 5, 6\}$ and $\lambda_{\rm up} \in \{1,2,3,4\}$ and selected the values yielding the best worst-group accuracy on the dev set.
Since no dev set is provided for MultiNLI, we tuned the hyperparameters on FEVER and applied them to MultiNLI.

\subsection{Results}\label{subsec:results}

We compared our proposed method (referred to as $\JTTm$, Eq.~(\ref{eqn:jtt-m})) against two baselines: $\ERM$ (Eq.~(\ref{eqn:erm})) and $\JTT$ (Eq.~(\ref{eqn:jtt})).
Table \ref{tab:main-results} shows the results for the average and worst-group performances of various approaches. 

As expected, $\ERM$ had the best average accuracy but performed poorly on the worst group across the two datasets. 
$\JTT$ and $\JTTm$ had improved performance on the worst group with slightly decreased average accuracies on both datasets compared with $\ERM$.
On FEVER, $\JTTm$ outperformed $\JTT$ in average accuracy while maintaining the same worst-group [SUP, neg] accuracy. 
On MultiNLI, $\JTTm$ performed significantly better on the worst group [Neut, neg] and maintained the same average accuracy as $\JTT$. 

We also observed larger variations in the results for FEVER. 
This is likely due to the smaller group sizes in FEVER. 
The worst group of MultiNLI [Neut, neg] accounted for around 3.5\% of the test set, while FEVER's [SUP, neg] was only 0.5\% of the test set and was about 5 times lower than the smallest group in MultiNLI in absolute numbers. 
For the same reason, another minority group of FEVER, [NEI, neg], also displayed a higher variation.

In addition, $\JTTm$ slightly reduced training time due to the smaller training set. 
Our Mahalanobis distance method detected 2,077 and 1,821 examples as outliers in the FEVER and MultiNLI error sets.
By eliminating these examples, we could reduce the training time while achieving results similar to or better than $\JTT$.

\subsection{Discussion}\label{subsec:discussion}

\begin{table}[t]
    \centering
    \begin{subtable}[ht]{1.0\columnwidth}
        \setlength{\tabcolsep}{10pt} 
        \begin{center}
        \begin{tabu}{lcc} 
        \toprule
          Group & $\Jtt$ & $\Jttm$$\;\;$ \\
        \midrule
{} [REF, no neg] & 79.9\STD{0.5} & 80.7\STD{0.3}$\;\;$\\
{} [REF, neg] & 93.8\STD{0.6} & 96.2{\STD{0.6}}$^\ast$\\
{} [SUP, no neg] & 94.7\STD{0.2} & 94.5\STD{0.1}$\;\;$\\
{} [SUP, neg] & 50.5\STD{3.5} & 50.2\STD{2.8}$\;\;$\\
{} [NEI, no neg] & 82.5\STD{0.5} & 83.0\STD{0.3}$\;\;$\\
{} [NEI, neg] & 71.5\STD{0.9} & 72.1\STD{3.3}$\;\;$\\
        \bottomrule
        \end{tabu}
        \caption{FEVER}\label{sub-tab:group-accs-a}\bigskip
        \end{center}
    \end{subtable}
    \begin{subtable}[ht]{1.0\columnwidth}
        \setlength{\tabcolsep}{10pt} 
        \begin{center}
        \begin{tabu}{lcc} 
        \toprule
          Group & $\Jtt$ & $\Jttm$$\;\;$ \\
        \midrule
{} [Contr, no neg] & 82.8\STD{0.7} & 82.8\STD{1.0}$\;\;$\\
{} [Contr, neg] & 91.9\STD{0.1} & 91.8\STD{0.6}$\;\;$\\
{} [Ent, no neg] & 82.6\STD{0.2} & 82.2\STD{1.1}$\;\;$\\
{} [Ent, neg] & 79.5\STD{0.5} & 78.9\STD{1.9}$\;\;$\\
{} [Neut, no neg] & 81.2\STD{0.6} & 81.7\STD{0.8}$\;\;$\\
{} [Neut, neg] & 75.5\STD{1.5} & 77.3{\STD{0.4}}$^\ast$\\
        \bottomrule
        \end{tabu}
        \caption{MultiNLI}\label{sub-tab:group-accs-b}
        \end{center}
    \end{subtable}
\caption{Accuracies and standard deviations for each group on (a) FEVER and (b) MultiNLI. ``$\ast$'' indicates statistical significance (paired t-test, $p < 0.05$).}\label{tab:group-accs}
\end{table}

The improvements for the MultiNLI worst group agree with our hypothesis: removing outliers from the upweighted error set improves model performance. 
As seen in Table~\ref{tab:group-accs}, all other groups of MultiNLI were either not affected by the removal of outliers or showed insignificant changes. 
On the other hand, removing outliers from the FEVER error set seemed to have a larger effect on groups other than the worst group [SUP, neg], especially on [REF, neg] and [NEI, neg].

\begin{figure}[t]
\centering
\includegraphics[width=1.0\columnwidth]{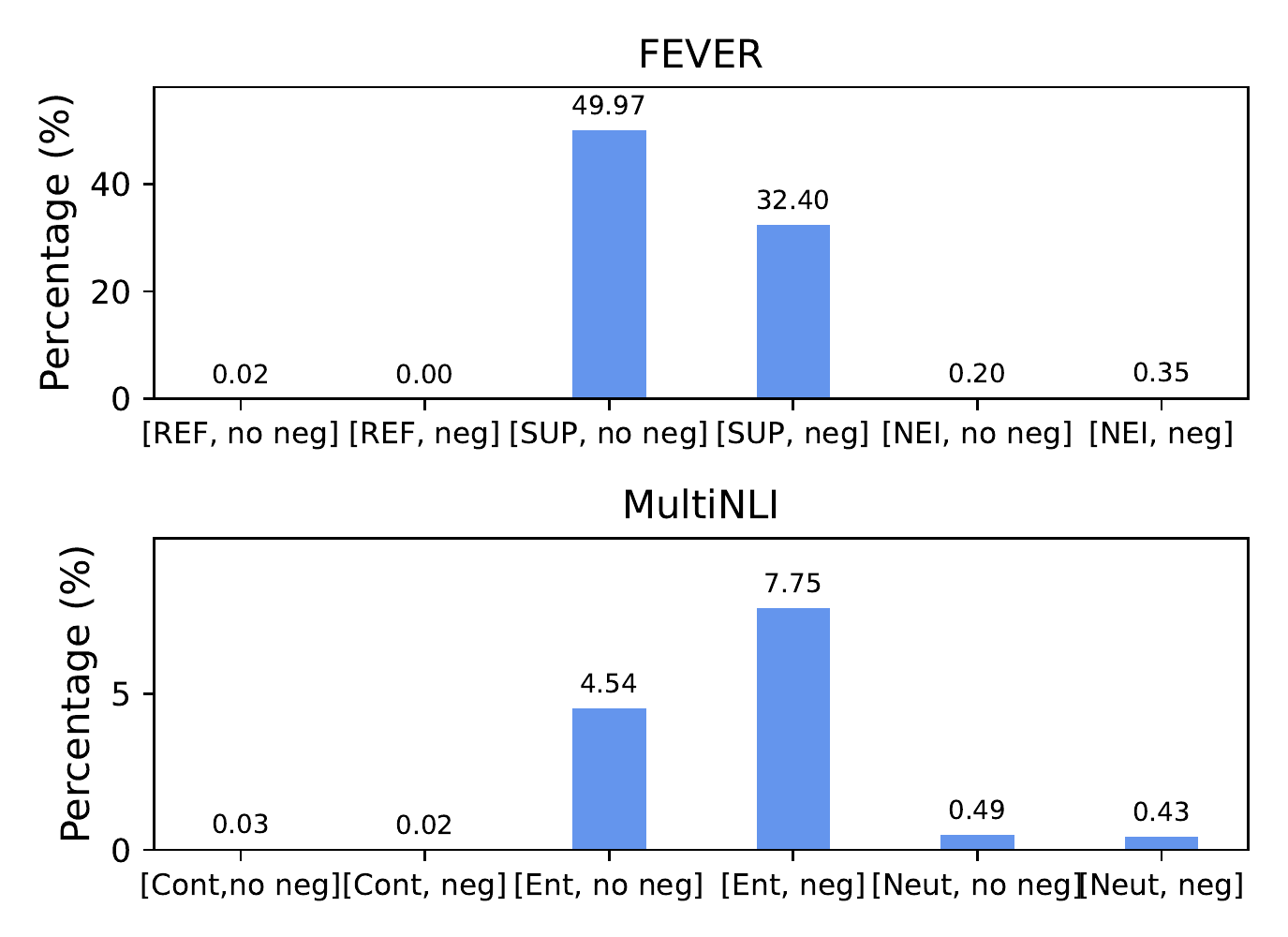}
\caption{
Percentage of OOD examples in the error set of each group. A large percentage of examples from classes SUP and Ent are regarded as outliers. FEVER's SUP has a much higher percentage removed compared with MultiNLI's Ent. All other groups contain only less than 1\% of examples regarded as outliers.
}
\label{fig:OOD-pcnt-in-group}
\end{figure}

We examined the group-wise percentage of the error-set OOD examples (i.e., the ones removed in $\JTTm$) to see how each group may be affected by the removal of their OOD examples (Figure \ref{fig:OOD-pcnt-in-group}).
Despite the improvements in groups [REF, neg] and [Neut, neg], few to no examples from these groups were regarded as outliers by the Mahalanobis distance method.
Instead, groups of classes SUP and Ent, whose performance does not improve when outliers are removed, contained the highest percentage of OOD examples. 
This suggests that these outliers can affect the model's decision boundaries among classes.

To investigate the properties of the OOD examples detected, we randomly sampled 100 examples from $S_{\rm in}$ and $S_{\rm out}$ for both FEVER and MultiNLI. 
For FEVER, we found 24 annotation errors in $S_{\rm out}$, much higher than the 1 annotation error in $S_{\rm in}$. 
For MultiNLI, $S_{\rm out}$ contained 10 annotation errors, whereas $S_{\rm in}$ contained 4. We show a sample of the annotation errors found in Table \ref{tab:annot-errors}.
This suggests that (1) the Mahalanobis distance method can detect at least a subset of annotation errors as outliers, and (2) the improvements in either the group or the overall performance may be partially due to the removal of these annotation errors.

\begin{table}[t]
\small
    \centering
    \begin{subtable}[ht]{1.0\columnwidth}
        \setlength{\tabcolsep}{4pt} 
        \begin{center}
        \begin{tabular}{lp{0.7\columnwidth}} 
        \toprule
        \textbf{Claim:} & {Nice \& Slow was released in \underline{1968}.} \\
        \textbf{Evidence:} 
        & {"Nice \& Slow" is a \underline{1998} single from Usher's second album My Way.} \\
        \multicolumn{2}{l}{\textbf{Annotated label:} \textsc{Supports}} \\
        \multicolumn{2}{l}{\textbf{Predicted label:} \textsc{Refutes}} \\
        \bottomrule
        \end{tabular}
        \end{center}

        \caption{FEVER}\label{sub-tab:annot-errors-a}\bigskip
    \end{subtable}
    
    \begin{subtable}[ht]{1.0\columnwidth}
        \setlength{\tabcolsep}{4pt} 
        \begin{center}
        \begin{tabular} {lp{0.7\columnwidth}}
        \toprule
        \textbf{Premise:} 
        & {So far, \underline{no promising treatments exist} according to Larry Gentilello.} \\
        \textbf{Hypothesis:} 
        & {Larry Gentilello asserted that \underline{effective} \underline{treatments already exist}, not just treatments that hold promise.} \\
        \multicolumn{2}{l}{\textbf{Annotated label:} Entailment} \\
        \multicolumn{2}{l}{\textbf{Predicted label:} Contradiction}\\
        \bottomrule
        \end{tabular}
        \end{center}
        \caption{MultiNLI}\label{sub-tab:annot-errors-b}
    \end{subtable}
\caption{Example of annotation errors from (a) FEVER and (b) MultiNLI. }\label{tab:annot-errors}
\end{table}

\section{Conclusion}

We have shown that the $\Jtt$ algorithm can benefit from pruning the error set before upweighting and training a second time, improving worst-group accuracy or overall accuracy on two popular datasets. 
We also showed that annotation errors may occur in the error set, hampering $\Jtt$'s effectiveness. 
These annotation errors can be mitigated by detecting and removing them with our Mahalanobis distance method. 
Investigating the effects of using other OOD-detection methods and finding a more effective way to tune the additional hyperparameters are directions for our future work.

\section*{Acknowledgments}

This work is supported by JST CREST Grants (JPMJCR18A6 and JPMJCR20D3) and MEXT KAKENHI Grants (21H04906), Japan.

\bibliography{anthology,custom}
\bibliographystyle{acl_natbib}

\end{document}